\documentclass[9pt,twocolumn, oneside]{osajnl}

\journal{josaa} 

\setboolean{shortarticle}{false}

\ifthenelse{\boolean{shortarticle}}{\colorlet{color2}{color2b}}{\colorlet{color2}{color2}} 

\usepackage{color}
\usepackage{bbm}
\usepackage{mathrsfs}
\usepackage{amsfonts}
\usepackage{amssymb}
\usepackage{graphicx}
\usepackage{array}
\usepackage{subfigure}
\usepackage{multirow}
\usepackage{amsmath}
\usepackage{bm}
\usepackage{upgreek}
\usepackage{float}
\newcommand{\format}[1]{\text{\bf #1}}

\newcommand{\tabincell}[2]{\renewcommand\arraystretch{1.0}\begin{tabular}{@{}#1@{}}#2\end{tabular}}
\usepackage{threeparttable}

\title{Experimental comparison of single-pixel imaging algorithms}

\author[1]{Liheng Bian}
\author[1]{Jinli Suo}
\author[1]{Qionghai Dai}
\author[1,2,*]{Feng Chen}
\affil[1]{Department of Automation, Tsinghua University, Beijing 100084, China}
\affil[2]{Beijing Key Laboratory of Security in Big Data Processing and Application, Beijing 100084, China}
\affil[*]{Corresponding author: chenfeng@mail.tsinghua.edu.cn}
\dates{Compiled \today}

\ociscodes{(170.3010) Image reconstruction techniques; (110.1758) Computational imaging; (150.1135) Algorithm.}

\doi{\url{http://dx.doi.org/10.1364/ao.XX.XXXXXX}}

\begin{abstract}
Single-pixel imaging (SPI) is a novel technique capturing 2D images using a photodiode, instead of conventional 2D array sensors. SPI owns high signal-to-noise ratio, wide spectrum range, low cost, and robustness to light scattering. Various algorithms have been proposed for SPI reconstruction, including the linear correlation methods, the alternating projection method (AP), and the compressive sensing based methods. However, there has been no comprehensive review discussing respective advantages, which is important for SPI's further applications and development. In this paper, we reviewed and compared these algorithms in a unified reconstruction framework. Besides, we proposed two other SPI algorithms including a conjugate gradient descent based method (CGD) and a Poisson maximum likelihood based method. Both simulations and experiments validate the following conclusions: to obtain comparable reconstruction accuracy, the compressive sensing based total variation regularization method (TV) requires the least measurements and consumes the least running time for small-scale reconstruction; the CGD and AP methods run fastest in large-scale cases; {the TV and AP methods are the most robust to measurement noise}. In a word, there are trade-offs between capture efficiency, computational complexity and robustness to noise among different SPI algorithms. We have released our source code for non-commercial use.
\end{abstract}

\setboolean{displaycopyright}{true}

\begin{document}

\maketitle
\thispagestyle{fancy}

\ifthenelse{\boolean{shortarticle}}{\ifthenelse{\boolean{singlecolumn}}{\abscontentformatted}{\abscontent}}{}

\section{Introduction}\label{sec:Introduction}

\begin{figure*}[!t]
\centering
\centerline{\includegraphics[width=0.8\textwidth]{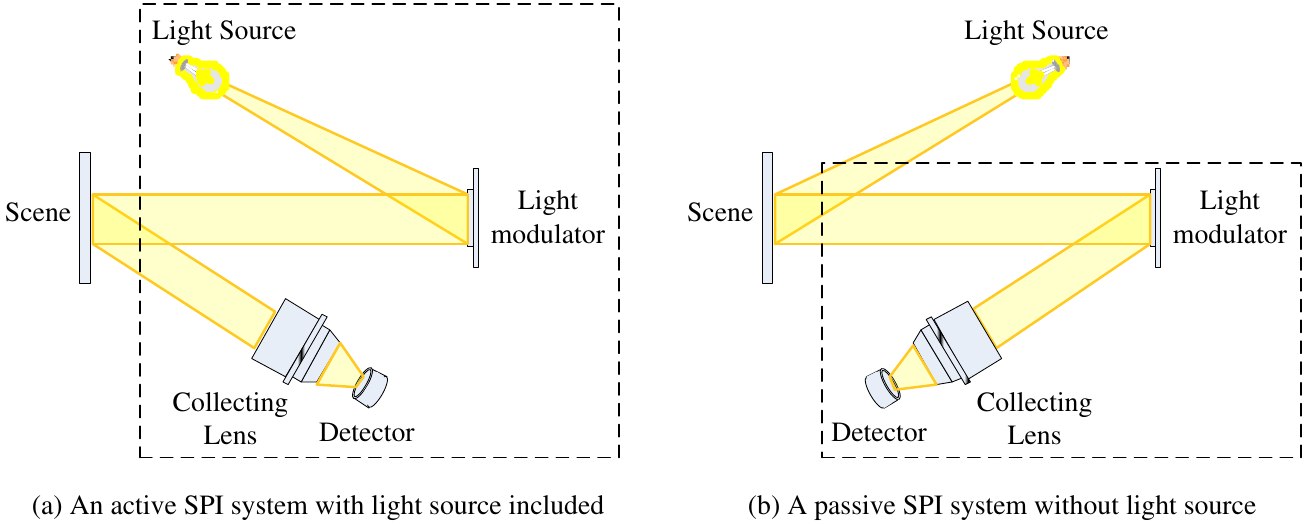}}
\caption{Light paths of two common single-pixel imaging systems. (a) In an active SPI system, the light modulator is placed between the active light source and the target scene. (b) In a passive SPI system, the light modulator is placed between the target scene and the detection module. This configuration doesn't need active light source included.}
\label{fig:fig_lightPath}
\end{figure*}

Single-pixel imaging (SPI) \cite{duarte2008single} is a novel imaging technique producing 2D images with a photodiode instead of conventional 2D array sensors. Specifically, SPI uses a light modulator such as a diffuser \cite{guo2016image} or a programmable spatial light modulator (SLM) to modulate light patterns. The correlated light from the target scene is finally collected by a photodiode, as shown in Fig. \ref{fig:fig_lightPath}.
The 2D target scene can be reconstructed from the modulation patterns and corresponding 1D measurements, using various algorithms including linear correlation methods \cite{bromberg2009ghost, gong2010a, ferri2010differential, sun2012normalized}, alternating projection method \cite{guo2016image} and compressive sensing (CS) based techniques \cite{katz2009compressive, compressive_scientificreport}.
Due to its high signal-to-noise ratio (SNR), wide spectrum range, low cost, flexible system configuration and robustness to light scattering, SPI has drawn more and more attentions in the last decade, and has been widely applied in multispectral imaging \cite{2016_SR_MSPI, wang2016hyperspectral, li2017efficient}, 3D modeling \cite{sun20133d, sun2016single}, optical encryption \cite{clemente2010optical, chen2013ghost}, remote sensing \cite{zhao2012ghost, gong2016three}, object tracking \cite{magana2013compressive, li2014ghost, gibson2017real}, imaging through atmospheric turbulence \cite{cheng2009ghost, zhang2010correlated}, etc.

SPI shares a similar imaging scheme to ghost imaging (GI) \cite{pittman1995optical, strekalov1995observation}. They both multiplex scene information into 1D measurements optically, and demultiplex the 2D target image computationally. GI originates from quantum optics, and produces scene images using the spatial correlation of entangled photon pairs (one interacts with the scene and the other not). Later, GI has been demonstrated to be applicable for classical thermal light \cite{bennink2002two, ferri2005high}. To get rid of recording modulation patterns, Shapiro \cite{shapiro2008computational} proposed using an SLM to modulate light, which coincides with the developments of SPI.



Although SPI and GI have attracted more and more attentions from various fields, their studies were conducted separately in computer science and optics. It is necessary to compare the reconstruction algorithms under the same framework. Such a unified perspective offers researchers an easy choice of  appropriate algorithms for their experiments. This could further push forward the development and applications of SPI.

In this paper, we reviewed SPI and GI algorithms in a unified reconstruction framework. Also, we proposed and tested two other methods for SPI reconstruction, including a conjugate gradient descent based method and a Poisson maximum likelihood based method. These algorithms are applied on both simulated data and real captured data, under different experiment settings including sampling ratio, image size, and noise level. Given comparable reconstruction accuracy, we investigate and compare their capture efficiency, computational complexity, and robustness to measurement noise.

The remainder of this paper is organized as follows: modeling and derivation of different SPI reconstruction algorithms are conducted in Sec.~\ref{sec:Theory}. The simulations and experiments are presented in Sec.~\ref{sec:Experiments}. Finally, we conclude this paper in Sec.~\ref{sec:Conclusions}.

\section{SPI algorithms}\label{sec:Theory}

The single-pixel imaging scheme is a linear system. Specifically, the measurement formation can be described as
\begin{eqnarray}\label{eqs:Formation_0}
\format A\format x &=& \format b,
\end{eqnarray}
where $\format A \in \mathbb {R}^{m\times n}$ denotes the light modulation matrix ($m$ modulation patterns, and each pattern consists of $n$ pixels), $\format x \in \mathbb {R}^{n\times 1}$ stands for the target scene (aligned as a vector) to be reconstructed, and $\format b \in \mathbb {R}^{m\times 1}$ is the measurement vector. The SPI reconstruction is to calculate $\format x$ from the modulation patterns $\format A$ and corresponding measurements $\format b$. In the following, we present the derivations of different algorithms, which are classified into three categories including non-iterative methods, linear iterative methods and non-linear iterative methods.

\subsection{Non-iterative methods}

The non-iterative methods perform direct reconstruction of the target scene without iteration. {One common and intuitive method is using inverse matrix to reverse the linear model. However, in most cases, there are usually $m$ measurements and $n$ signals where $m\neq n$, which means that the light modulation matrix $\format A$ is non-symmetric. To make the problem solvable, $\format A^T$ is multiplied to both sides of Eq. (\ref{eqs:Formation_0}), and the model becomes $\format A^T\format A\format x = \format A^T\format b$. Then $\format x$ is calculated as $\format x = (\format A^T\format A)^{-1}\format A^T\format b$. According to ref. \cite{7590036}, this is essentially equivalent to minimizing $||\format A\format x - \format b||_{l_2}$ when $m\geqslant n$. However, $\format A^T\format A$ is non-full rank when $m<n$, and the method would not work.}

Based on the fact that the SPI measurements stand for the correlation between modulation patterns and the target scene, $\format x$ can be reconstructed by correlating modulation patterns with corresponding measurements as
\begin{eqnarray}
\label{eqs:alg_TGI}
\format x &=& \frac{1}{m}\sum_{i=1}^{m}\left({b_{i}}-\left\{ b_i \right\}\right)\format a_i\\ \nonumber
&=& \left\{ b_i\format a_i \right\} - \left\{ b_i \right\}\left\{ \format a_i \right\},
\end{eqnarray}
where $b_i$ is the $i$th measurement in $\format b$, $\format a_i$ is the $i$th modulation pattern (row) in $\format A$, and $\left\{ \cdot \right\}$ is the ensemble average in terms of $i$ defined as $\left\{ b_i \right\}=\frac{1}{m}\sum_{i=1}^{m}b_i$ and $\left\{ \format a_i \right\}=\frac{1}{m}\sum_{i=1}^{m}\format a_i$. Basically, the reconstruction is a weighted summation of modulation patterns, with weights being corresponding measurements. A larger measurement means the modulation pattern is more similar to the target scene, and thus leads to stronger correlation and a larger weight of the pattern in reconstruction. Therefore, this kind of methods are also referred to linear correlation methods.
Statistically, if the intensity of each pixel over different patterns is independently identically distributed (i.i.d.), the average of all the patterns approaches uniform as their number goes to infinity, which could produce a high-quality reconstruction.

Some variant algorithms have been proposed to improve the correlation based reconstruction, among which the differential ghost imaging (DGI) method \cite{ferri2010differential, gong2010a} is widely used. DGI takes illumination fluctuations into account, and introduces an additional detector to detect the total intensity of each illumination pattern. Specifically, if we denote the pattern intensity vector as $\format s \in \mathbb {R}^{m\times 1}$, DGI is performed as
\begin{eqnarray}
\label{eqs:alg_DGI}
\format x &=& \left\{ b_i\format a_i \right\} - \frac{\left\{b_i \right\}}{\left\{ s_i \right\}}\left\{ s_i\format a_i \right\}.
\end{eqnarray}
Intuitively, DGI uses the light's total intensity to normalize illumination patterns, which can improve the signal-to-noise ratio (SNR) of final reconstruction. Due to its robustness, we use DGI to represent the non-iterative algorithms in following simulations and experiments.

\subsection{Linear iterative methods}

\subsubsection{Gradient descent (GD)}

The SPI reconstruction can be treated as an error (between real measurements and its estimates) reduction process. Mathematically, it can be formulated as a quadratic minimization problem:
\begin{eqnarray}\label{eqs:gd1}
\min ~~~ L(\format x) &=& ||\format A\format x - \format b||_{l_2}^2,
\end{eqnarray}
where the $l_2~norm$ is defined as $||\format x||_{l_2} = \sqrt{\sum_{i}(x_i)^2}$. The gradient of the above objective function is derived as
\begin{eqnarray}\label{eqs:gd2}
\format p &=& \frac{\partial L(\format x)}{\partial \format x} \\ \nonumber
&=& 2\format A^T(\format A\format x - \format b),
\end{eqnarray}
Using the gradient, one can reconstruct the target scene by iteratively updating $\format x$ as
\begin{eqnarray}\label{eqs:gd3}
\format x' &=& \format x - \Delta_{\format x}\format p,
\end{eqnarray}
where $\Delta_{\format x}$ denotes the step size. Usually, the updating process converges when the objective function becomes smaller than a given threshold.

To converge fast, the optimum step size is needed, which can be calculated by solving the following problem:
\begin{eqnarray*}\label{eqs:gd4}
\min ~~~ L(\Delta_{\format x}) &=& ||\format A(\format x - \Delta_{\format x}\format p) - \format b||_{l_2}^2 \\ \nonumber
&=& ||(\format A\format x - \format b) - \Delta_{\format x}\format A\format p||_{l_2}^2.
\end{eqnarray*}
The gradient of the above function to $\Delta_{\format x}$ is
\begin{eqnarray*}\label{eqs:gd5}
\frac{\partial L(\Delta_{\format x})}{\partial \Delta_{\format x}} &=& -2(\format A\format p)^T\left[(\format A\format x - \format b) - \Delta_{\format x}\format A\format p) \right] \\ \nonumber
&=& 2(\format p^T\format A^T\format r + \Delta_{\format x}\format p^T\format A^T\format A\format p),
\end{eqnarray*}
where $\format r = \format b - \format A\format x$ is the residual error vector. Assigning $\frac{\partial L(\Delta_{\format x})}{\partial \Delta_{\format x}} = 0$ yields the closed-form solution of optimum $\Delta_{\format x}$ as
\begin{eqnarray}\label{eqs:gd6}
\Delta_{\format x} &=& -\frac{\format p^{T}\format A^T\format r}{\format p^{T}\format A^T\format A\format p}.
\end{eqnarray}

\subsubsection{Conjugate gradient descent (CGD)}

The conjugate gradient descent method (CGD) is also designed to solve the quadratic minimization problem in Eq. (\ref{eqs:gd1}), but with the requirement that $\format A$ need to be symmetric and positive-definite \cite{hestenes1952methods}. Similar to the matrix inversion method, researchers multiply $\format A^T$ to both sides of Eq. (\ref{eqs:Formation_0}) and get
\begin{eqnarray}\label{eqs:cgd1}
\format A^T\format A\format x = \format A^T\format b ~~~\Longleftrightarrow~~~  \format A'\format x = \format b',
\end{eqnarray}
where $\format A' \in \mathbb{R}^{n\times n}=\format A^T\format A$ and $\format b' \in \mathbb{R}^{n\times 1}=\format A^T\format b$. The residual error vector is $\format r^k = \format b' - \format A'\format x^k = \format A^T\format b - \format A^T\format A\format x$, and the gradient is defined as
\begin{eqnarray}\label{eqs:cgd2}
\format p^{(k)} &=& - \format r^{(k-1)} - \frac{\format r^{(k-1)T}\format r^{(k-1)}}{\format r^{(k-2)T} \format r^{(k-2)}}\format p^{(k-1)},
\end{eqnarray}
where $k$ in the superscript denotes the $k$th iteration. The step size is set to be
\begin{eqnarray}\label{eqs:cgd3}
\Delta_\format x^{(k)} &=& \frac{\format r^{(k-1)T}\format r^{(k-1)}}{\format p^{(k)T}\format A'\format p^{(k)}}.
\end{eqnarray}

According to extensive studies \cite{hestenes1952methods, luenberger1973introduction}, CGD converges faster than the conventional gradient descent method. This owes to that the designed gradient in each iteration are conjugate to each other. Theoretically, CGD converges within no more than $n$ iterations for the reconstruction of an $n$-pixel image.

\subsubsection{Poisson maximum likelihood}

Utilizing photons' statistic that they arrive at a sensor following the Poisson distribution \cite{bian2016fourier}, we propose to apply the maximum likelihood estimation method \cite{akaike1998information} for SPI reconstruction. This method aims to estimate $\format x$ by maximizing the likelihood of producing the measurements $b_i \in \format b$. The objective function of the Poisson maximum likelihood algorithm is
\begin{eqnarray}\label{eqs:poisson1}
\max ~~~ \prod_{i=1}^{m} \frac{e^{-(\format a_i\format x)}(\format a_i\format x)^{b_i}}{b_i!},
\end{eqnarray}
which is equivalent to
\begin{eqnarray}\label{eqs:poisson2}
\min ~~~ L(\format x) &=& -\log \prod_{i=1}^{m} \frac{e^{-(\format a_i\format x)}(\format a_i\format x)^{b_i}}{b_i!} \\ \nonumber
\Leftrightarrow ~~~ \min ~~~ L(\format x) &=& \sum_{i=1}^{m}\left(\format a_i\format x - b_i\log(\format a_i\format x)\right).
\end{eqnarray}
The gradient can be derived as
\begin{eqnarray}\label{eqs:poisson3}
\format p &=& \frac{\partial L(\format x)}{\partial \format x} \\ \nonumber
&=& \sum_{i=1}^{m}\left(\format a_i^T - \frac{b_i}{\format a_i\format x}\format a_i^T\right) \\ \nonumber
&=& \format A^T \left(\frac{\format A\format x - \format b}{\format A\format x}\right).
\end{eqnarray}
{The target scene is reconstructed following the same gradient descent updating in Eq. (\ref{eqs:gd3}).} Since the gradient in Eq. (\ref{eqs:poisson3}) is non-linear, it is hard to derive a closed-form solution of the optimum step size $\Delta_\format x$. Hence, we use the standard backtracking line search method \cite{luenberger1973introduction, boyd2004convex} to calculate the step size in each iteration. The search process is summarized in Alg. (\ref{alg:step}).


\begin{algorithm}
\caption{Backtracking line search method}\label{alg:step}
\begin{algorithmic}[1]
\Procedure{Input}{$\format p, \alpha \in [0.01,0.3], \beta \in [0.1,0.8]$}
\State $\Delta_{\format{x}} \gets 1$
\While{$~~~L(\format x + \Delta_{\format{x}}\format p) > L(\format x) - \alpha\Delta_{\format{x}}\format p^T\format p~~~$}
\State $\Delta_{\format{x}}\gets \beta\Delta_{\format{x}}$
\EndWhile\label{euclidendwhile}
\State \textbf{return} $\Delta_{\format{x}}$
\EndProcedure
\end{algorithmic}
\end{algorithm}

\subsubsection{Alternating projection}

The alternating projection (AP) method for SPI reconstruction was proposed by Guo et al. \cite{guo2016image}, which is adapted from the alternating projection method for phase retrieval \cite{Phase_Comparison, FPM_Nature}. This method considers the reconstruction from the view of spatial spectrum. It treats the SPI measurement $b_i \in \format b$ as the zero-spatial-frequency coefficient of the light field arriving at the photodiode, which is denoted as $\format l_{\format a_i} = \format a_i^T \odot \format x$ with $\odot$ standing for dot product. {In each iteration, AP switches between the Fourier and spatial domains to add support constraints.
In Fourier space, $b_i$ is the support constraint for zero spatial frequency, and stands for the total light intensity of $\format l_{\format a_i}$. Therefore, $\format l_{\format a_i}$ is updated as
\begin{eqnarray*}
\format l_{\format a_i}' &=& \format l_{\format a_i}\frac{b_i}{\format a_i\format x}.
\end{eqnarray*}
In spatial space, the modulation patterns are the support constraints, and the target image $\format x$ is updated as
\begin{eqnarray*}
\format x' &=& \format x + \frac{\format a_i^T}{\max(\format a_i)^2}(\format l_{\format a_i}'-\format l_{\format a_i}),
\end{eqnarray*}
Incorporating the above calculations produces the updating principle of AP as
\begin{eqnarray}\label{eqs:ap}
\format x' &=& \format x - \\\nonumber
&&\frac{\format a_i^T\odot(\format a_i^T\odot\format x)}{\max(\format a_i)^2}\frac{\format a_i\format x - b_i}{\format a_i\format x}.
\end{eqnarray}
In each iteration, all the measurements are sequentially used in Eq. (\ref{eqs:ap}) to update the target image.}

\subsection{Non-linear iterative methods}

Compressive sensing \cite{donoho2006compressed, candes2006stable, cande2008introduction} aims to reconstruct signals from underdetermined linear systems by introducing signal priors, and it can be utilized for SPI reconstruction to reduce measurements.
There are two widely used priors for natural images, including the sparse representation prior and the total variation (TV) regularization prior. The first one states that natural images are sparse when represented by some overcomplete or orthogonal bases, such as the discrete cosine transform {(DCT)} basis used for the JPEG compression standard \cite{duarte2008single}. The TV regularization prior states that the gradient's integral of a natural image is statistically low. {There also exist other compressive sensing studies for SPI with different extensions \cite{m2013compressive, Yu14, gong2015high, hu2015patch}. However, most of them fall into these two kinds of priors, which therefore will be mainly discussed in this article.}

\subsubsection{Sparse representation prior}\label{sec:Sparse}

Mathematically, the basis transform matrix and corresponding coefficient vector are respectively denoted as $\format D$ and $\format c$, and the optimization model becomes
\begin{eqnarray}\label{eqs:cs1}
\min &~& ||\format c||_{l_0} \\ \nonumber
s.t. &~& \format D\format x = \format c\\ \nonumber
&~& \format A\format x = \format b,
\end{eqnarray}
where the $l_0$ norm calculates the number of non-zero entries in $\format c$, and describes its sparsity.

{Usually, ones use $l_1$ norm to approximate $l_0$ norm for easier reconstruction, and the above optimization can be conducted under a gradient descent framework, using the augmented Lagrange multiplier (ALM) method \cite{ALM_YiMa, NIPS2011_4434}. ALM has been proved to be a robust and fast algorithm for $l_1$ minimization \cite{yang2010fast}.}
By introducing a Lagrange multiplier $\format y$ to incorporate the equality constraints into the objective function, our goal is to minimize the Lagrangian function of Eq. (\ref{eqs:cs1}) as
\begin{eqnarray}\label{eqs:cs2}
\min ~ L
\hspace*{-6pt} &=& \hspace*{-6pt} ||\format c||_{l_1} + <\format y_1, \format D\format x - \format c> + \frac{\mu_1}{2}||\format D\format x - \format c ||_{l_2}^2 \\ \nonumber
&~& ~~~~~~~ + <\format y_2, \format A\format x - \format b> + \frac{\mu_2}{2}||\format A\format x - \format b||_{l_2}^2 \\ \nonumber
\Leftrightarrow \min ~ L \hspace*{-6pt} & = & \hspace*{-6pt} ||\format c||_{l_1} + \frac{\mu_1}{2}||\format D\format x - \format c + \frac{\format y_1}{\mu_1}||_{l_2}^2 \\\nonumber
&& + \frac{\mu_2}{2}||\format A\format x - \format b + \frac{\format y_2}{\mu_2}||_{l_2}^2,
\end{eqnarray}
where $<\cdot>$ stands for the inner product, and $\mu_{\{1, 2\}}$ are the parameters balancing different optimization items.

The variables in the above objective function include $\format c$,  $\format x$, $\format y_1$, $\format y_2$, $\mu_1$ and $\mu_2$. Following the iterative scheme of ALM, the updating principle of each variable is to minimize the Lagrangian function while keeping the other variables constant. The detailed derivations are as follows.

{\vspace{3mm} \noindent{\bf Optimize $\format c$.~~~}} Removing all the items irrelevant to $\format c$, the objective function becomes
\begin{eqnarray}\label{eqs:csc1}
\min ~~~ L(\format c) &=& ||\format c||_{l_1} + \frac{\mu_1}{2}||\format D\format x - \format c + \frac{\format y_1}{\mu_1}||_{l_2}^2.
\end{eqnarray}
Following the ALM algorithm, the updating rule of $\format c$ is
\begin{eqnarray}\label{eqs:csc2}
\format c &=& \mathbb {T}_{\frac{1}{\mu_1}}\left(\format D\format x + \frac{\format y_1}{\mu_1}\right),
\end{eqnarray}
where $\mathbb {T}_{\frac{1}{\mu_1}}\left(\cdot\right)$ is a thresholding operator defined as
\begin{equation*}\label{eqs:csc2}
\mathbb {T}_{\frac{1}{\mu_1}}(x) = \begin{cases}
x - \frac{1}{\mu_1}, & x > \frac{1}{\mu_1}\\
x + \frac{1}{\mu_1}, & x < -\frac{1}{\mu_1}\\
0,& others.
\end{cases}
\end{equation*}

{\vspace{3mm} \noindent{\bf Optimize $\format x$.~~~}} Removing all the items irrelevant to $\format x$ yields
\begin{eqnarray}\label{eqs:csx1}
\min ~~~ L(\format x) &=& \frac{\mu_1}{2}||\format D\format x - \format c + \frac{\format y_1}{\mu_1}||_{l_2}^2 \\\nonumber
&&+ \frac{\mu_2}{2}||\format A\format x - \format b + \frac{\format y_2}{\mu_2}||_{l_2}^2,
\end{eqnarray}
and the gradient is
\begin{eqnarray*}\label{eqs:csx2}
\frac{\partial L(\format x)}{\partial \format x} &=& \mu_1\format D^T(\format D\format x - \format c + \frac{\format y_1}{\mu_1}) + \mu_2\format A^T(\format A\format x - \format b + \frac{\format y_2}{\mu_2}).
\end{eqnarray*}
By setting $\frac{\partial L(\format x)}{\partial \format x} = \format 0$, we can easily obtain the closed-form solution of $\format x$ as
\begin{eqnarray}\label{eqs:csx3}
\format x &=& (\mu_1\format D^T\format D + \mu_2\format A^T\format A)^{-1} \\\nonumber
&&\left[\mu_1\format D^T(\format c - \frac{\format y_1}{\mu_1}) + \mu_2\format A^T(\format b - \frac{\format y_2}{\mu_2})\right].
\end{eqnarray}

{\vspace{3mm} \noindent{\bf Optimize $\format y$ and $\mu$.~~~}} In the ALM algorithm, the Lagrange multipliers $\format y$ and the balancing parameter $\mu$ are updated as
\begin{eqnarray}\label{eqs:csy1}
\format y_1' &=& \format y_1 + \mu_1(\format D\format x - \format c) \\ \nonumber
\format y_2' &=& \format y_2 + \mu_2(\format A\format x - \format b),
\end{eqnarray}

\begin{eqnarray}\label{eqs:csmu1}
 u_1' &=& \min(\rho  u_1, \mu_{1max})  \\ \nonumber
 u_2' &=& \min(\rho  u_2, \mu_{2max}),
\end{eqnarray}
{where $\rho$ and $\mu_{max}$ are the parameters set by users to adjust the growing speed and maximum of $\mu$.}

\begin{table*}[!t]
\centering
\caption{Comparison of reconstruction principles among different SPI algorithms}\label{tab:algorithms}
\renewcommand\arraystretch{2.5} 
\begin{threeparttable}
\centerline{
\begin{tabular}{c| c| c| c| c| c}
\cline{1-6}
\multicolumn{2}{c|}{Methods}& Perspective & \multicolumn{3}{c}{Principles}\\
\cline{1-6}
\multirow{3}*{ \tabincell{c}{No \\ iteration} } & \tabincell{c}{Matrix \\ inversion} & \tabincell{c}{Formation \\ fitting} & \multicolumn{3}{c}{$\format x = (\format A^T\format A)^{-1}\format A^T\format b$}  \\
\cline{2-6}
 & \tabincell{c}{Conventional \\ correlation\cite{bromberg2009ghost}} & \multirow{2}*{ \tabincell{c}{Measurement: \\ scene-pattern \\ correlation} } & \multicolumn{3}{c}{Eq. (\ref{eqs:alg_TGI}): ~~~ $\left\{ b_i\format a_i \right\} - \left\{ b_i \right\}\left\{ \format a_i \right\}$}  \\
\cline{2-2}\cline{4-6}
 & DGI \cite{ferri2010differential} & &  \multicolumn{3}{c}{Eq. (\ref{eqs:alg_DGI}): ~~~ $\left\{ b_i\format a_i \right\} - \frac{\left\{ b_i \right\}}{\left\{ s_i \right\}}\left\{ s_i\format a_i \right\}$} \\
\cline{1-6}
\multirow{5}*{ \tabincell{c}{Linear \\ iteration} } & \multicolumn{2}{c|}{} & Problem description &  Gradient $\format p$ & Step $\Delta_\format x$ \\
\cline{2-6}
 & \tabincell{c}{ \tabincell{c}{Gradient \\ descent} } & \multirow{2}*{ \tabincell{c}{Formation \\ fitting} } & \tabincell{c}{Eq. (\ref{eqs:gd1}): \\ $\min ~ ||\format A\format x - \format b||_{l_2}^2$}  &  \tabincell{c}{Eq. (\ref{eqs:gd2}): \\ $\format A^T(\format A\format x - \format b)$}   &  \tabincell{c}{Eq. (\ref{eqs:gd6}): \\ $-\frac{\format p^{T}\format A^T\format r}{\format p^{T}\format A^T\format A\format p}$}  \\
\cline{2-2}\cline{4-6}
 & \tabincell{c}{Conjugate \\ gradient \\ descent} & & \tabincell{c}{Eq. (\ref{eqs:cgd1}): \\ $\format A^T\format A\format x = \format A^T\format b$} & \tabincell{c}{Eq. (\ref{eqs:cgd2}): \\ $- \format r^{(k-1)} - \frac{\format r^{(k-1)T}\format r^{(k-1)}}{\format r^{(k-2)T} \format r^{(k-2)}}\format p$ }   & \tabincell{c}{Eq. (\ref{eqs:cgd3}): \\ $\frac{\format r^{T}\format r}{\format p^{T}\format A^T\format A\format p}$ }  \\
\cline{2-6}
 & \tabincell{c}{Poisson \\ maximum \\ likelihood} & \tabincell{c}{Signal \\ statistic}   & \tabincell{c}{Eq. (\ref{eqs:poisson2}): $\min$ \\ $\sum_{i=1}^{m}\left(\format a_i\format x - b_i \log(\format a_i\format x)\right)$ }  &  \tabincell{c}{Eq. (\ref{eqs:poisson3}): \\ $\format A^T\left(\frac{\format A\format x - \format b}{\format A\format x}\right)$}  & Alg. (\ref{alg:step}) \\
\cline{2-6}
 & \tabincell{c}{Alternating \\ projection \cite{guo2016image}} & \tabincell{c}{Measurement: \\ zero-spatial-frequency \\ coefficient} &  \tabincell{c}{Eq. (\ref{eqs:Formation_0}): \\ $\format A\format x = \format b$}   & \tabincell{c}{Eq. (\ref{eqs:ap}): \\ $\frac{\format a_i^T\odot(\format a_i^T\odot\format x)}{\max(\format a_i)^2}\frac{\format a_i\format x - b_i}{\format a_i\format x}$ }   &  1 \\
\cline{1-6}
\multirow{2}*{\tabincell{c}{Non-linear \\ iteration}} & \tabincell{c}{Sparse \\ representation \cite{duarte2008single} } & \multirow{2}*{ \tabincell{c}{Image \\ prior} } & \multirow{2}*{\tabincell{c}{Eq. (\ref{eqs:cs1}): \\ $\min ~ ||\format c||_{l_0}$ \\ $s.t. ~ \format D\format x = \format c, \format A\format x = \format b$} }     &  \multicolumn{2}{c}{\multirow{2}*{See Eqs. (\ref{eqs:cs2} - \ref{eqs:csmu1}) in Sec. \ref{sec:Sparse}}} \\
\cline{2-2}
& \tabincell{c}{Total \\ variation} & &  &  \multicolumn{2}{c}{} \\
\cline{1-6}
\end{tabular}
}
\end{threeparttable}
\end{table*}

\subsubsection{Total variation regularization prior}\label{sec:TV}

As described in ref. \cite{Suo2016130}, the gradient of an image $\format x$ can be denoted as $\format c = \format G\format b$ where $\format G$ is the gradient calculation matrix. Here we use $l_1$ norm to calculate the image's total variation, namely the gradient's integral, and the optimization model becomes
\begin{eqnarray}\label{eqs:tv1}
\min &~& ||\format c||_{l_1} \\ \nonumber
s.t. &~& \format G\format x = \format c\\ \nonumber
&~& \format A\format x = \format b.
\end{eqnarray}
We can see that the optimization shares the same form as the sparse representation based reconstruction in Eq. (\ref{eqs:cs1}), and we can utilize the same algorithm introduced in Sec. \ref{sec:Sparse} to solve the TV problem.

\vspace{5mm}
On the above, we introduced various SPI reconstruction algorithms. A summarization of their optimization models and reconstruction principles are listed in Tab. \ref{tab:algorithms}.

\section{Quantitative comparison}\label{sec:Experiments}

In this section, we compare the performance of the above SPI reconstruction algorithms on both simulated and experimental data, to show their pros and cons.

\subsection{Settings}\label{sec:settings}

We first clarify several experiment settings. (1) For quantitative comparison, we use the normalized root-mean-square error (RMSE) as the metric. Normalized RMSE is defined as {$\sqrt{\frac{\{(\format I_1 - \format I_2)^2\}}{\{\format I_1\}}}$}, with $\{\cdot\}$ being the ensemble average operation as defined before. It measures the relative difference between the ground-truth image $\format I_1$ and the reconstructed image $\format I_2$.
(2) For the initialization of $\format x$ in the above iterative algorithms, one can theoretically set any value. Here we set $\format x^{(0)} = \format 0$ for all the algorithms for a fair comparison.
(3) The iteration of iterative algorithms usually stops until either it reaches a preset number, or the objective function becomes smaller than a given threshold. Here, we use the change of residual error between two subsequent iterations as the criterion, namely {$||\format r||^{(k)} - ||\format r||^{(k-1)}$} where $\format r = \format b - \format A\format x$. If it is small enough (we set the threshold as $10^{-2}$), the iteration stops. Also, we set the maximum iteration number as three times of the pixel number. To ensure convergence, we set the minimum iteration number as 30.

\begin{figure}[!b]
\centering
\centerline{\includegraphics[width=0.5\textwidth]{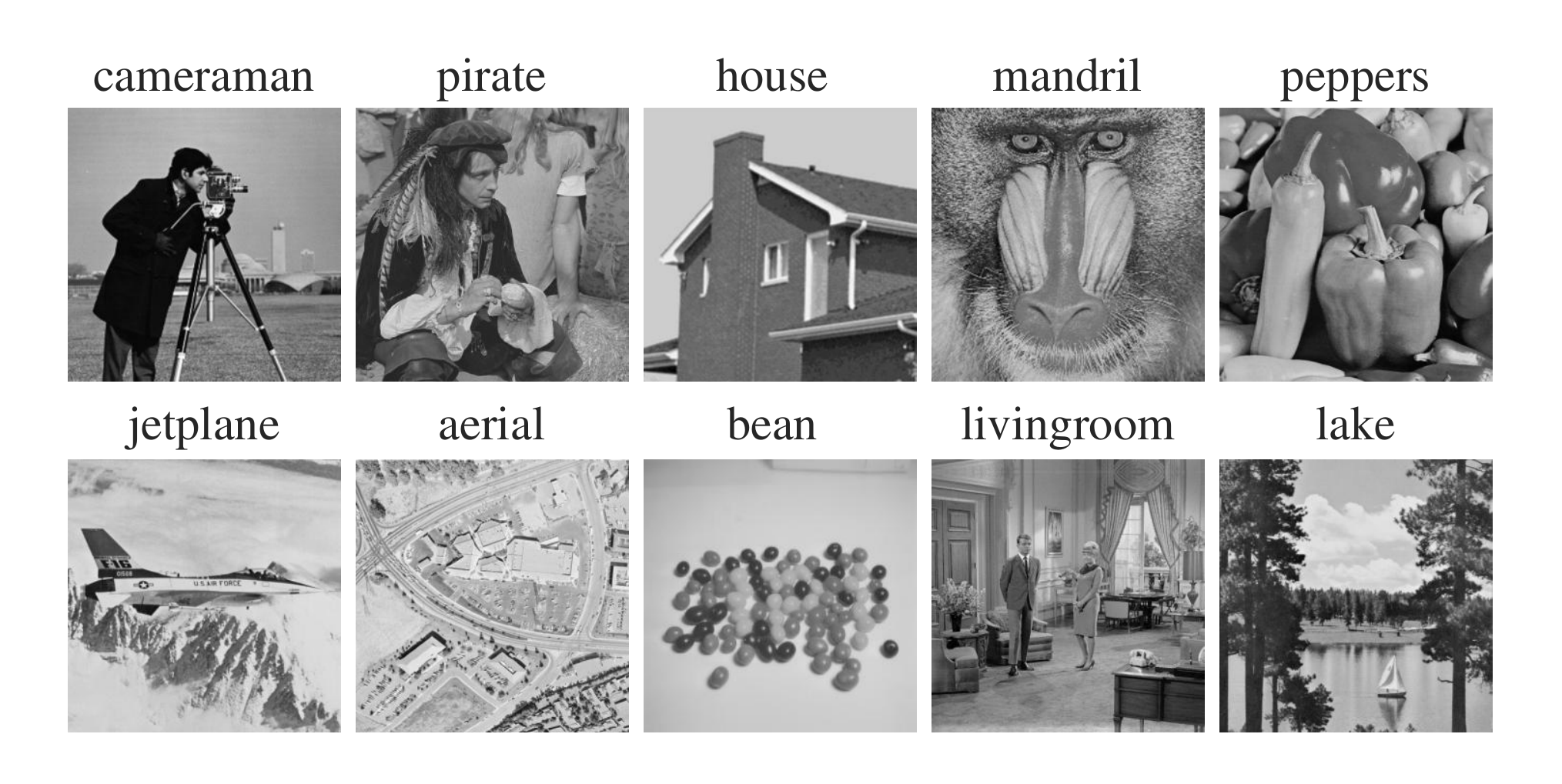}}
\caption{Images used in the simulations as target scenes.}
\label{fig:fig_images}
\end{figure}

\begin{figure*}[!t]
\centering
\centerline{\includegraphics[width=0.85\textwidth]{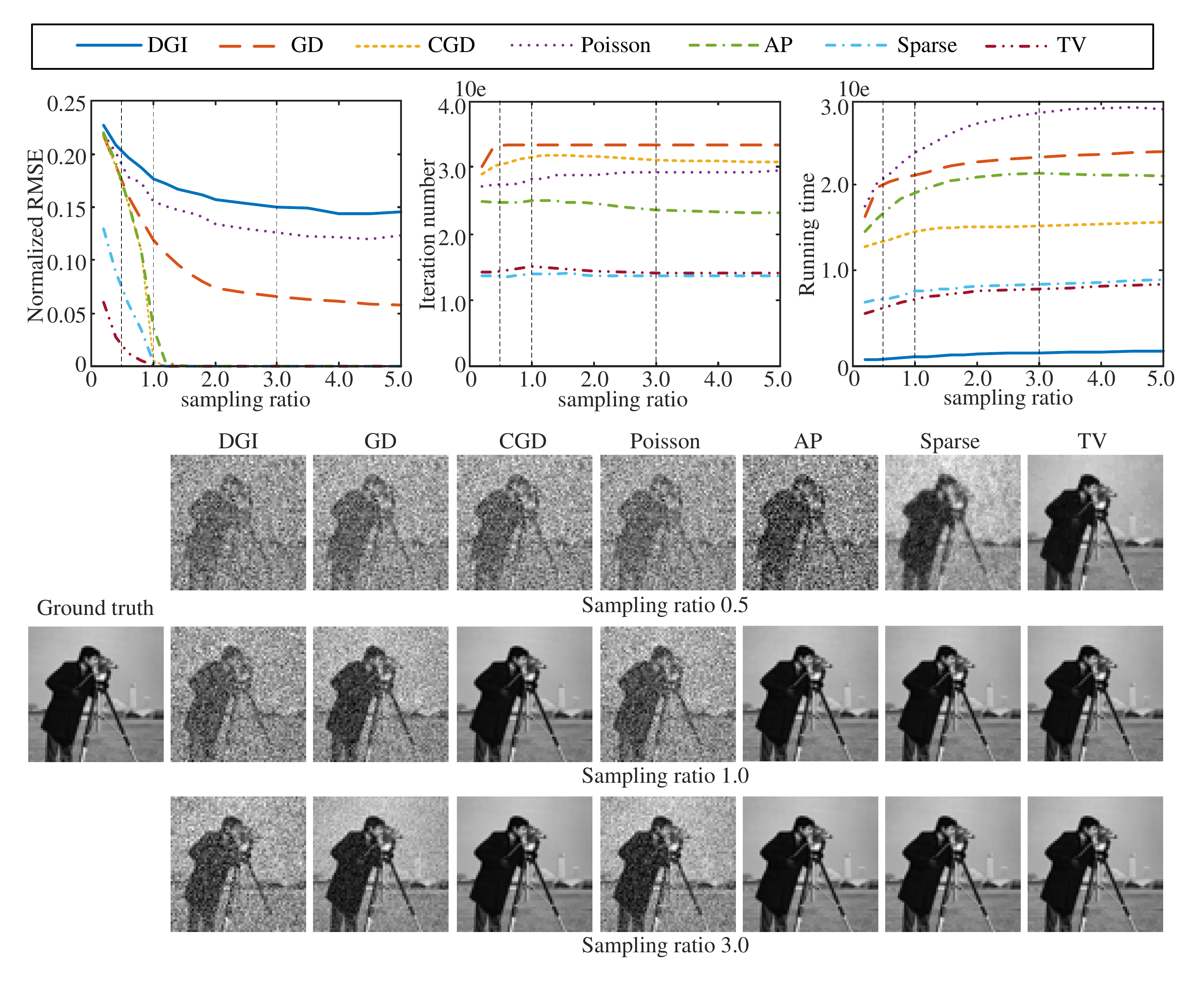}}
\caption{SPI results of different algorithms under different sample ratios. The three curve graphs respectively show reconstruction error, iteration number and running time. The vertical coordinates of iteration number and running time are exponential to make the figures clearer. Note that there is no iteration number of the DGI method because it is non-iterative. Besides, exemplar reconstructed images under the sampling ratios of 0.5, 1.0 and 3.0 are shown on the bottom.}
\label{fig:fig_sampleRatio}
\end{figure*}

\subsection{Simulations}\label{sec:Simulations}

In the following simulations, we set 10 widely used "Standard" test images \cite{Gonzalez2006, imageset, imageset2} as target scenes {(the pixel values are normalized to [0, 1])}, as shown in Fig. \ref{fig:fig_images}. For each algorithm, the reconstruction RMSEs on these images are averaged to quantitatively characterize its performance. Note that we only present the reconstructed images of 'cameraman' in the following figures to save space. The widely used random modulation is applied to synthesize measurements based on the formation model in Eq. (\ref{eqs:Formation_0}). There are two parameters we would like to clarify, including sampling ratio and image size. Sampling ratio is defined as the ratio between the number of measurements and that of the signals (pixels) to be reconstructed. It determines capture efficiency. Except for the simulation specifically discussing sampling ratio, it is set constant as 1. Image size is the pixel number of the final reconstructed images, which is also the same as that of illumination patterns. Similarly, image size is set constant as $64 \times 64$ pixels in the following simulations except for the subsection specifically discussing it. Besides, we repeat each simulation for 20 times, and average the evaluations to produce final quantitative results. All the algorithms are implemented using Matlab on an Intel Core i7-4790 3.6 GHz CPU computer, with 16G RAM and 64 bit Windows 7 system.

\subsubsection{Sampling ratio}\label{sec:sampleRatio}

\begin{figure*}[!t]
\centering
\centerline{\includegraphics[width=0.85\textwidth]{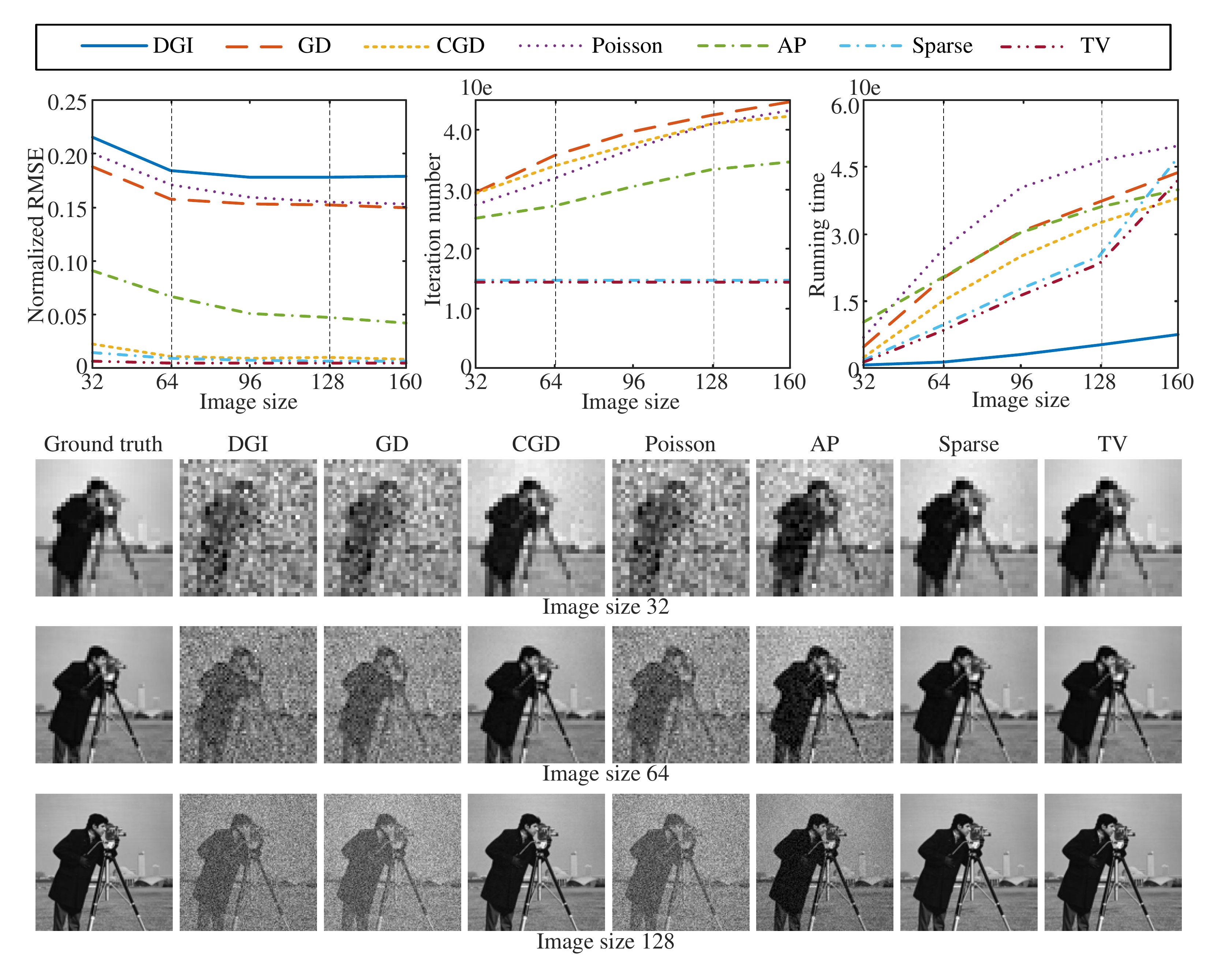}}
\caption{SPI reconstructions of different algorithms under different reconstructed image sizes. The upper figures show the reconstruction error, iteration number and running time of different algorithms. The exemplar recosntructed images of $32 \times 32$, $64 \times 64$ and $128 \times 128$ pixels are shown on the bottom.}
\label{fig:fig_imageSize}
\end{figure*}

We test the algorithms on simulated data of different sampling ratios ranging from 0.2 to 5. The reconstruction results are shown in Fig. \ref{fig:fig_sampleRatio}, from which we have the following observations:
\begin{itemize}
  \item Reconstruction error largely decreases as sampling ratio increases to 1. When sampling ratio continues to increase, the error decreasing speed becomes slower.
  \item Iteration number and running time do not increase much as sampling ratio increases. {Though, one should note that the capture time increases as sampling ratio becomes larger.}
  \item The conventional linear correlation method and the two linear iterative methods including Poisson maximum likelihood based method and the GD method produce large reconstruction error even with the sampling ratio being 5. They own low capture efficiency.
  \item The non-linear iterative methods (especially the TV method) need much less measurements than the other methods for comparable reconstruction quality, and converges faster than most of the linear iterative methods except for the CGD method. As a demonstration, the TV method needs only 50$\%$ measurements of signals to obtain normalized RMSE smaller than 0.03. This benefits from the introduced image priors that provide extra scene information.
\end{itemize}

\subsubsection{Image size}\label{sec:imageSize}

Image size is another important factor affecting both reconstruction quality and computational complexity.
Here we set sampling ratio constant as 1, and run the algorithms on simulated data of different image sizes from 32$\times$32 pixels to 160$\times$160 pixels. The results are presented in Fig. {\ref{fig:fig_imageSize}}, from which we can see that large image size leads to less reconstruction error. However, large image size takes more iterations and running time. Besides, we have the following observations:
\begin{itemize}
  \item The running time of the non-linear iterative methods increases much faster than the other methods as image size increases. This dues to the non-linear calculations and updating of multiple introduced variables.
  \item As image size increases, the iteration number of the non-linear iterative methods does not increase. This means that if sampling ratio is high enough, a small number of iterations is enough for the non-linear methods to converge, which might be attributed to the introduced image priors.
  \item DGI, CGD and AP take the least running time when image size excceeds 160 $\times$ 160 pixels. However, DGI produces much error in final recosntruction. Therefore, CGD and AP outperform the other methods in large-scale SPI reconstruction considering both reconstruction quality and computational complexity.
\end{itemize}

\subsubsection{Noise}\label{sec:noise}

\begin{figure*}[!t]
\centering
\centerline{\includegraphics[width=0.83\textwidth]{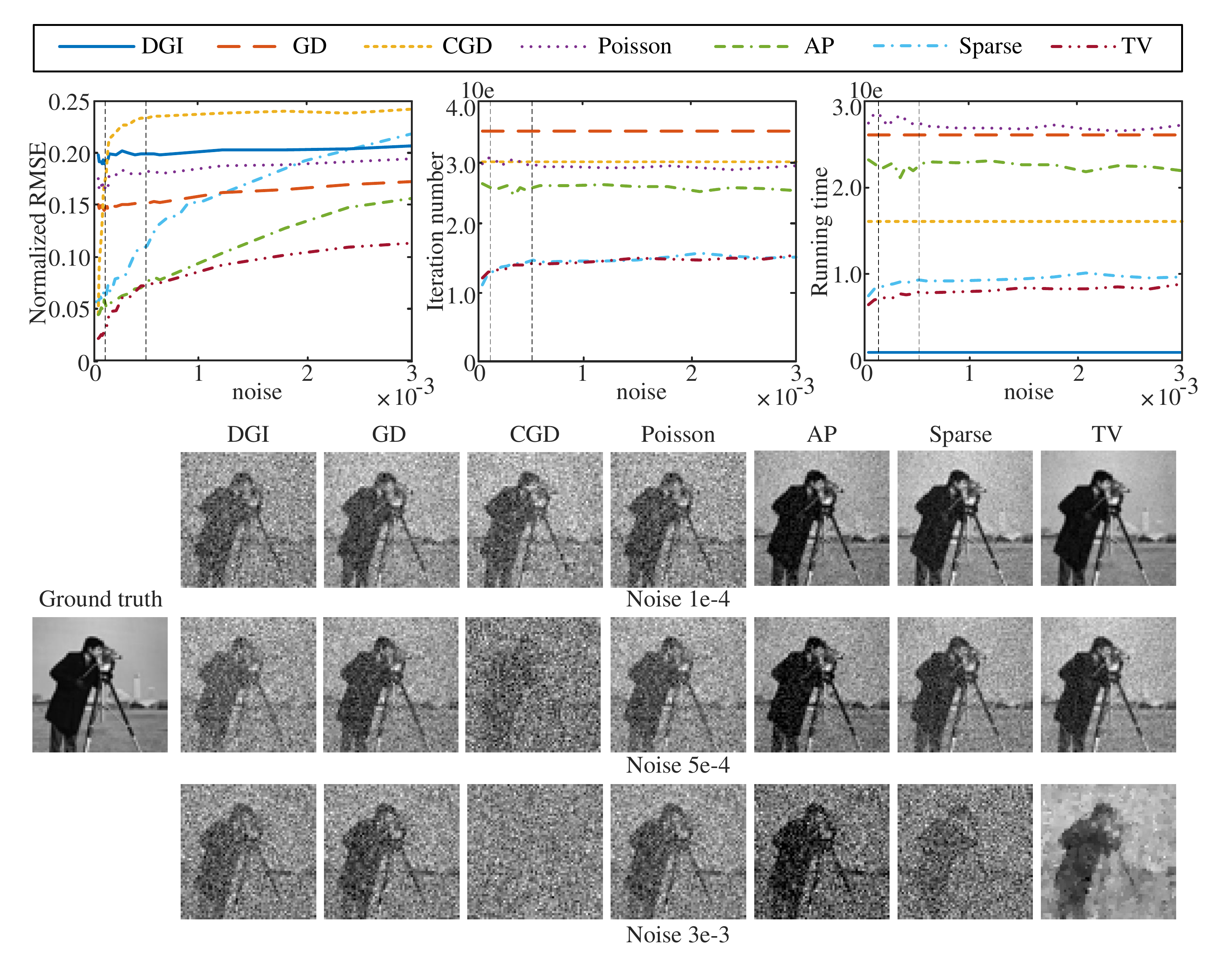}}
\caption{SPI results of different algorithms under different measurement noise levels. Quantitative reconstruction error, iteration number and running time are curved in the three upper figures. Exemplar reconstructed images under the noise level (the ratio between noise's standard deviation and pixel number) of 1e-4, 5e-4 and 3e-3 are shown on the bottom.}
\label{fig:fig_GaussianNoise}
\end{figure*}

In the above simulations, we assume no measurement noise. However, measurements in reality are always corrupted with noise arising from various causes such as ambient light and circuit current.
In this subsection, we study the influence of measurement noise on final reconstruction and test the algorithms' robustness. Here we assume Gaussian white noise following the probability distribution
\begin{eqnarray}\label{eqs:noise}
P(n) = \frac{1}{\sqrt{2\pi}\sigma}\exp\left(-\frac{n^2 }{2\sigma^2}\right),
\end{eqnarray}
where $n$ stands for noise, and $\sigma$ is its standard deviation (std). {Noise level is characterized by the ratio between $\sigma$ and pixel number. We vary noise level by changing the ratio from 0 to 3e-3.} Sampling ratio is 1, and image size is $64\times 64$ pixels. Corresponding reconstruction results are shown in Fig. \ref{fig:fig_GaussianNoise}, from which we can draw the following conclusions:
\begin{itemize}
  \item As measurement noise increases, all the reconstructions degrade.
  \item The CGD and compressive sensing based sparse representation algorithms degrade faster than the other methods.
  \item The TV and AP methods are the two most robust methods to attenuate measurement noise.
\end{itemize}

\subsection{Experiment}\label{sec:realExperiments}

\begin{figure*}[!t]
\centering
\centerline{\includegraphics[width=0.8\textwidth]{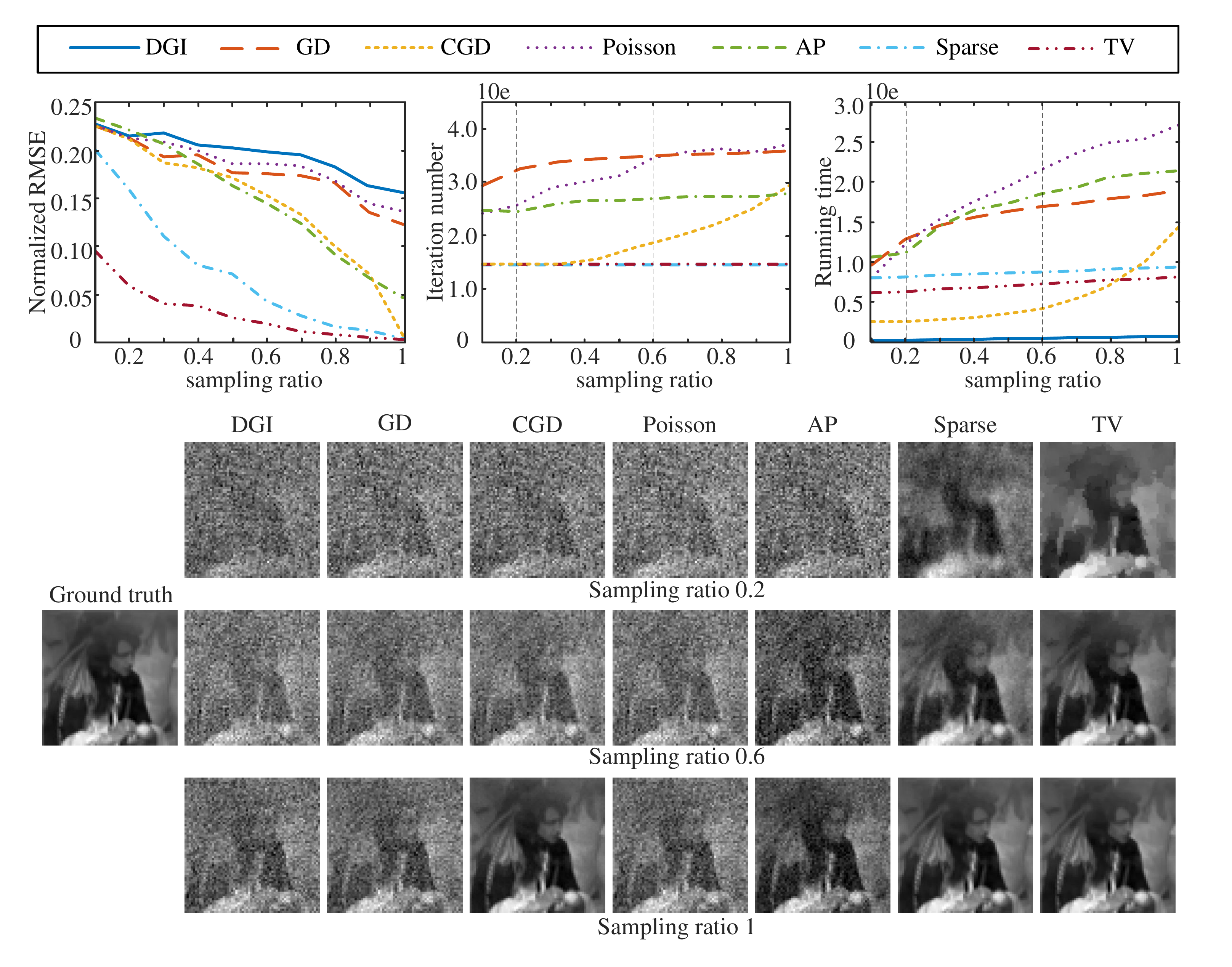}}
\caption{Experimental SPI results of different algorithms. The first row shows reconstruction error, corresponding iteration number and running time under different sampling ratios (0.1 - 1). Also, exemplar reconstructed images under the sampling ratios of 0.2, 0.6 and 1 are shown on the bottom.}
\label{fig:fig_real}
\end{figure*}

\begin{figure}[!t]
\centering
\centerline{\includegraphics[width=0.46\textwidth]{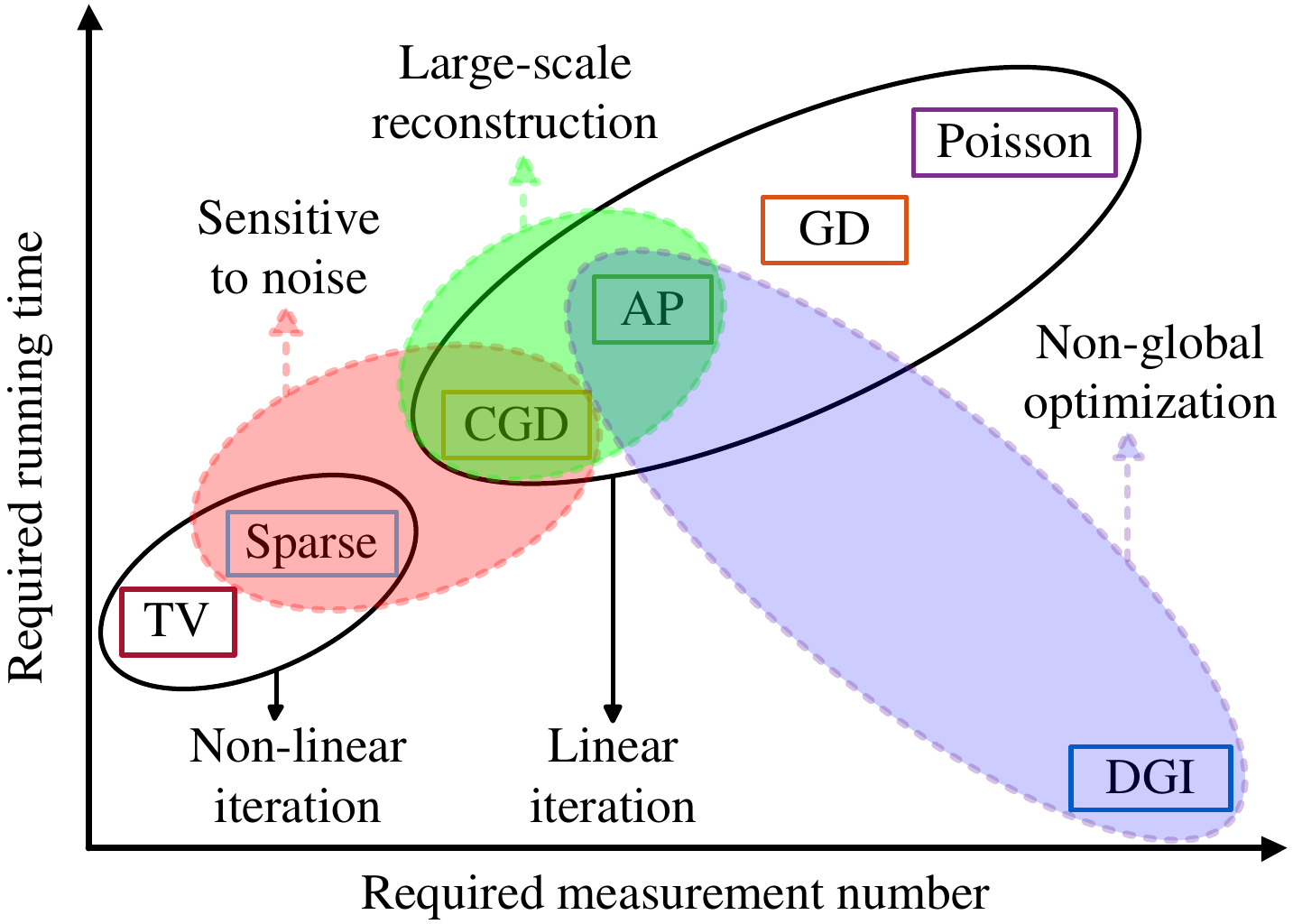}}
\caption{Comparison diagram of different SPI algorithms. The horizon axis states required measurement number for comparable reconstruction quality (i.e., capture efficiency), and the vertical axis shows required running time (i.e., reconstruction efficiency). The dotted red circle contains the sparse representation and CGD algorithms that are sensitive to measurement noise. The dotted purple circle contains the two non-global optimization algorithms including the AP and DGI methods. The dotted green circle contains the AP and CGD algorithms that take the least running time for large-scale reconstruction.}
\label{fig:fig_choice}
\end{figure}

To further compare the performance of different SPI algorithms, we built an SPI setup to capture real data. Similar to the reported SPI setup in ref. \cite{bian2016efficient}, we used a commercial projector's illumination module (numerical aperture of the projector lens is 0.27) and a digital micromirror device (DMD, Texas Instrument DLP Discovery 4100 Development Kit, .7XGA) for light modulation. Illumination patterns of 64 $\times$ 64 pixels are sequentially projected onto a printed transmissive film (34 mm $\times$ 34 mm) as the target scene (the "pirate" image in Fig. \ref{fig:fig_images}). The correlated light is recorded by a high-speed photodiode (Thorlabs DET100 Silicon photodiode, 340--1100 nm) together with a 14-bit acquisition board (ART PCI8514). The sampling rate is set as 10kHz. We utilize the self-synchronization technique \cite{SUO201565} to synchronize the DMD and the detector. Based on the observation in Sec. \ref{sec:sampleRatio} that reconstruction error of most algorithms is smaller than 0.05 as sampling ratio reaches 1, we set sampling ratio changing from 0.1 to 1, and utilize the reconstruction of the TV method under the sampling ratio of 3 as ground truth. The reconstructed results are presented in Fig. \ref{fig:fig_real}, from which we can see that the experiment results arrive at the same conclusions as those of the above simulations.

\section{Conclusions}\label{sec:Conclusions}

In this article, we reviewed various single-pixel imaging algorithms in a unified reconstruction framework, including the differential ghost imaging method (DGI), the gradient descent method (GD), the alternating projection method (AP), the sparse representation method and the total variation regularization method (TV). Also, we proposed two other methods for SPI reconstruction, including the conjugate gradient descent method (CGD) and the Poisson maximum likelihood method. These algorithms perform SPI reconstruction from different perspectives. The DGI method considers measurements as the correlation between target scene and illumination patterns, which indicates their similarities. Both the GD and CGD methods treat SPI reconstruction as a formation fitting process. The Poisson maximum likelihood method utilizes the statistic that random individual photons arrive at sensors following the Poisson distribution. The AP method considers the reconstruction from the view of spatial spectrum. It treats SPI measurements as the zero-spatial-frequency coefficients of the light fields arriving at detector. The two compressive sensing based methods introduce natural images' priors into optimization, and focus on under-determined SPI reconstruction with measurements less than signals.

These algorithms can be classified into three categories according to their iteration type, including the non-iterative methods (DGI), the linear iterative methods (GD, CGD, Poisson maximum likelihood method, AP), and the non-linear iterative methods (sparse representation method, TV). Generally, the non-iterative methods need the least running time but the most measurements. The non-linear iterative methods needs the least measurements. A visual comparison of the algorithms is presented in Fig. \ref{fig:fig_choice}.

These algorithms can also be classified into global optimization and non-global optimization methods, in terms of whether using all the measurements simultaneously in each update \cite{yeh2015experimental}. The DGI algorithm and the AP algorithm belong to non-global methods, while the other methods methods are global. Because non-global methods don't need to store all the patterns and measurements in each update, the DGI and AP methods own high storage efficiency.

We test and compare these algorithms on both simulated data and real captured data under different parameter settings including sampling ratio, image size and measurement noise. Based on the studies, we conclude that (1) considering capture efficiency, the TV method needs the least measurements for comparable reconstruction quality; (2) In terms of computational complexity, the TV and CGD methods take the least time for small-scale reconstruction; In large-scale cases, the CGD and AP methods run fastest; (3) {For robustness to measurement noise, the TV and AP methods are the most robust}. In a word, there are trade-offs between capture efficiency, computational complexity and noise robustness among different SPI reconstruction algorithms. One can choose appropriate algorithms according to specific SPI configurations and applications. Our open source code of all the algorithms can be downloaded at http://www.sites.google.com/site/lihengbian.

%

%
%
%
%

This work was funded by the National Natural Science Foundation of China (61327902, 61671266), the National High-tech Research and Development Plan (2015AA042306), and the Research Project of Tsinghua University (20161080084).


\ifthenelse{\equal{\journalref}{ol}}{%
\clearpage
\bibliographyfullrefs{sample}
}{}


\ifthenelse{\equal{\journalref}{aop}}{%
\section*{Author Biographies}
\begingroup
\setlength\intextsep{0pt}
\begin{minipage}[t][6.3cm][t]{1.0\textwidth} 
  \begin{wrapfigure}{L}{0.25\textwidth}
    \includegraphics[width=0.25\textwidth]{john_smith.eps}
  \end{wrapfigure}
  \noindent
  {\bfseries John Smith} received his BSc (Mathematics) in 2000 from The University of Maryland. His research interests include lasers and optics.
\end{minipage}
\begin{minipage}{1.0\textwidth}
  \begin{wrapfigure}{L}{0.25\textwidth}
    \includegraphics[width=0.25\textwidth]{alice_smith.eps}
  \end{wrapfigure}
  \noindent
  {\bfseries Alice Smith} also received her BSc (Mathematics) in 2000 from The University of Maryland. Her research interests also include lasers and optics.
\end{minipage}
\endgroup
}{}

\end{document}